\pdfoutput=1

\documentclass{article}
\usepackage{amsmath, amssymb}
\usepackage{enumitem}
\usepackage{hyperref}
\usepackage{adjustbox}

\usepackage[final]{acl}

\usepackage{times}
\usepackage{latexsym}

\usepackage[T1]{fontenc}

\usepackage[utf8]{inputenc}

\usepackage{microtype}

\usepackage{inconsolata}

\usepackage{graphicx}

\usepackage{enumerate}
\usepackage{amsmath}
\usepackage{multirow}
\title{Fuzzy Reasoning Chain (FRC): An Innovative Reasoning Framework from Fuzziness to Clarity}

\author{
Ping Chen$^{1,2}$\textsuperscript{\dag} \quad
Xiang Liu$^{1,2}$\textsuperscript{\dag} \quad
Zhaoxiang Liu$^{1,2}$\textsuperscript{*} \quad
Zezhou Chen$^{1,2}$ \quad 
Xingpeng Zhang$^{3}$ \\
\textbf{Huan Hu}$^{1,2}$ \quad
\textbf{Zipeng Wang}$^{1,2}$ \quad
\textbf{Kai Wang}$^{1,2}$ \quad
\textbf{Shuming Shi}$^{1,2}$ \quad
\textbf{Shiguo Lian}$^{1,2}$\textsuperscript{*}\\[2mm]
$^{1}$ Data Science \& AI Research Institute, China Unicom \\
$^{2}$ Unicom Data Intelligence, China Unicom \\
$^{3}$ School of Computer Science and Software Engineering, Southwest Petroleum University \\
\texttt{\{chenp181, liux750, liuzx178, liansg\}@chinaunicom.cn},
\texttt{xpzhang@swpu.edu.cn}
}

\begin{document}
\maketitle

\footnotetext[1]{Corresponding authors.}
\footnotetext[2]{Equal contribution.}
\begin{abstract}
With the rapid advancement of large language models (LLMs), natural language processing (NLP) has achieved remarkable progress. Nonetheless, significant challenges remain in handling texts with ambiguity, polysemy, or uncertainty. We introduce the Fuzzy Reasoning Chain (FRC) framework, which integrates LLM semantic priors with continuous fuzzy membership degrees, creating an explicit interaction between probability-based reasoning and fuzzy membership reasoning. This transition allows ambiguous inputs to be gradually transformed into clear and interpretable decisions while capturing conflicting or uncertain signals that traditional probability-based methods cannot. We validate FRC on sentiment analysis tasks, where both theoretical analysis and empirical results show that it ensures stable reasoning and facilitates knowledge transfer across different model scales. These findings indicate that FRC provides a general mechanism for managing subtle and ambiguous expressions with improved interpretability and robustness.
\end{abstract}

\section{Introduction}

With the rapid advancement of large-scale language models (LLMs) \cite{guo2025deepseek,Devlin2019BERT}, natural language processing (NLP) has achieved remarkable progress across various domains. However, reasoning with texts that exhibit ambiguity, polysemy, and uncertainty remains a significant challenge. In tasks such as sentiment analysis, ethical judgment, and security review, texts often contain complex emotions, such as "\textbf{\textit{Though dissatisfied, still acceptable.}}", which convey both negative sentiment and a degree of acceptance. This complexity makes traditional approaches, based on fixed labels or static rules, inadequate for capturing the multidimensional semantics \cite{pang2008opinion, bradley1994measuring}.

Current \textit{fuzzy reasoning} approaches quantify ambiguity through membership  degrees; however, their reliance on manually defined rules limits their adaptability to dynamic and evolving contexts \cite{taboada2011lexicon}. Although \textit{Chain-of-Thought} (CoT) reasoning enhances transparency, its discrete decision-making process often struggles with complex texts, such as those containing sarcasm and contradictions \cite{fei2023reasoning, wei2022chain, wang2023self}. Consequently, both fuzzy reasoning and CoT methods face limitations when addressing fuzzy and uncertain texts, underscoring the need for a more adaptive, stable, and transparent reasoning framework.

In this paper, we propose the \textit{Fuzzy Reasoning Chain} (FRC) framework to address challenges arising from ambiguous and uncertain text, as illustrated in Fig.~\ref{FRC}. FRC follows the standard step-by-step reasoning procedure typical of sentiment analysis and extends conventional chain-of-thought approaches by replacing discrete probability assignments with continuous fuzzy membership degrees. This transition from probability-based reasoning to fuzzy membership-based reasoning, which we refer to as the probability-to-membership collision, is the core methodological innovation. By capturing conflicting and ambiguous signals that probabilities alone cannot express, FRC enables a more nuanced and robust representation of sentiment while naturally identifying “Other” or unspecified categories. This fuzzy-to-clear reasoning paradigm ensures stability and transparency. We analyze the convergence behavior of FRC and corroborate its practical utility through comprehensive experiments, demonstrating its effectiveness on complex, fuzzy texts and potential for reasoning transfer across model scales.

The main contributions of FRC are as follows:

\begin{figure*}[ht]
    \centering
\includegraphics[width=0.98\linewidth]{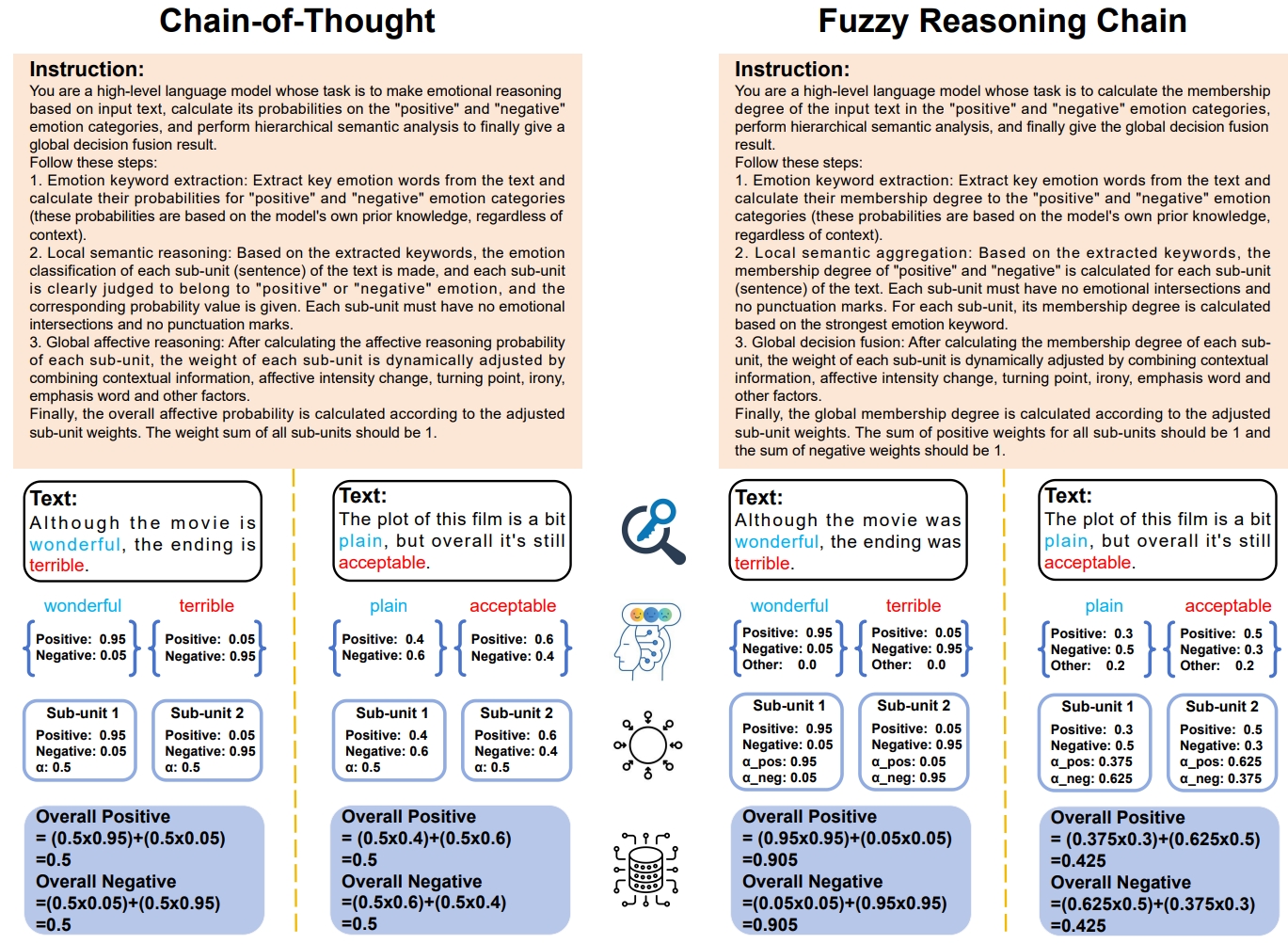}
    \caption{\textbf{Comparison of Chain-of-Thought (CoT, left) and Fuzzy Reasoning Chains (FRC, right) in sentiment analysis}. Both follow the same reasoning steps for fairness, reflecting typical sentiment analysis procedures and highlighting the interaction between probability-based reasoning and fuzzy membership. Unlike CoT, FRC membership degrees are unconstrained and do not need to sum to one, which allows any sentiment not explicitly covered in the prompt to be assigned to an “Other” category, capturing ambiguity and uncertainty that probability-based CoT cannot represent. FRC refines sentiment weights using fuzzy membership. In the first example, CoT shows balanced probabilities (0.5 each), while FRC captures strong conflicting sentiment (0.905 each). In the second example, CoT remains neutral (0.5 each), whereas FRC indicates a fuzzy sentiment state (0.425 each), enabling a transition from fuzzy to clear sentiment analysis.}
    \label{FRC}
\end{figure*}

\begin{itemize}
    \item \textbf{Core Methodological Innovation:}FRC integrates discrete probability reasoning with continuous fuzzy membership, enabling clearer interpretation of ambiguous inputs and capturing conflicting or uncertain signals that traditional probability-based methods cannot. This makes it the first framework to combine fuzzy membership with LLM reasoning in this way.
    \item \textbf{Convergence Analysis:} Ensures stable reasoning by analyzing robustness, monotonicity, and semantic completeness, supported by both theoretical analysis and experimental validation.
   \item \textbf{Empirical Validation and Generalization:} Experiments demonstrate that FRC enhances model performance on uncertain and ambiguous inputs, reflecting its generalization capacity and potential in lightweight knowledge transfer scenarios.
\end{itemize}

\section{Related Work}

\subsection{Sentiment Analysis}

Sentiment analysis is a fundamental task in NLP that aims to identify and classify the sentiment polarity of text \cite{pang2008opinion}. Research has explored sentiment dictionaries, machine learning, deep learning, and large model approaches. The use of emotion dictionaries enhances the original English  \cite{Whissell2009emotion} and Chinese emotion dictionary, thereby increasing its applicability to natural language samples. Support Vector Machines (SVM), Bayesian methods, ridge regression, and other classical machine learning techniques are also employed in natural language analysis \cite{Jemai2021ML, Hourrane2019Sentiment}. After the emergence of deep learning, it has become widely utilized in natural language sentiment analysis \cite{Zhang2018survey, Tang2015Document, Felbo2017Using}. Numerous methods employing Long Short-Term Memory (LSTM) networks and attention mechanisms have been proposed in this field. Regularized LSTM \cite{Qian2017Regularized} and Sentiment-Aware Bidirectional LSTM (SAB-LSTM) \cite{Kumar2020SAB} integrate linguistic resources, such as sentiment dictionaries, negative words, and intensity words, into the LSTM framework to more accurately capture emotional nuances in sentences. Attention mechanisms effectively capture the significance of each contextual word in relation to a specific target aspect \cite{Lei2016Rationalizing, Tiwari2022KEAHT}. The BERT \cite{Devlin2019BERT} model is a groundbreaking advancement in machine reading comprehension and can be applied across various fields, including sentiment analysis. 

In recent years, large language models (LLMs) have propelled advancements in unsupervised and self-supervised sentiment analysis, significantly enhancing performance on sentiment classification tasks \cite{Devlin2019BERT, yinhan2019roberta, Chen2024Emotion, Zhang2024Affective}. The literature \cite{Zhang2024Refashioning} emphasizes the importance of contextual information in enhancing the sentiment estimation of LLMs. Deep prompt tuning and low-rank adaptation effectively enhance the performance of large models in sentiment analysis \cite{Peng2024Customising}. Chain-of-Thought (CoT) are an effective strategy for enhancing the performance of LLMs and are also utilized for emotion classification \cite{fei2023reasoning, Duan2024Implicit}. In addition to CoT reasoning that relies solely on textual information \cite{fei2023reasoning}, audio and visual modalities are also taken into account to propose a cross-modal filtering and fusion (CMFF) module, which facilitates multi-modal sentiment analysis \cite{Li2024Multimodal}. The generative CoT strategy is employed for zero-shot and few-shot emotion classification tasks \cite{Gu2024AGCVT, Wu2025Few, Liu2024Zero}. However, these LLMs still predominantly depend on static labeling systems, making it difficult to handle complex sentiment conflicts and mixed emotions in texts effectively.

\subsection{Fuzzy Reasoning}

These sentiment analysis methods rely on discrete classification labels such as "positive", "negative", or "neutral", which struggle to address ambiguity, uncertainty, and mixed sentiment expressions \cite{bing2012sentiment}. To overcome these limitations, fuzzy reasoning approaches have been explored. 
Multi-dimensional sentiment models have been employed to capture the complex emotional components of text \cite{bradley1994measuring}. Fuzzy clustering simulates the continuous nature of sentiment distributions, providing a more nuanced approach to sentiment analysis \cite{peizhuang1983pattern}. Lexicon-based sentiment classifiers have been enhanced with fuzzy logic to create sentiment distributions more accurately reflect human cognitive processes in uncertain cases \cite{taboada2011lexicon}. There are also scholars who integrate the learning capabilities of deep learning with the uncertainty processing abilities of fuzzy logic to offer users more accurate emotional predictions \cite{Chaturvedi2019Fuzzy, Sweidan2022Word, Do2024Multimodal, Wang2025Exploring}. The emotion enhancement inference model integrates word embedding, an emotion dictionary, and fuzzy inference \cite{Yan2022Emotion}. However, mainstream approaches still struggle with transparency in the reasoning process and lack dynamic adjustment of sentiment labels during inference.

\subsection{Chain-of-Thought (CoT)}

CoT reasoning has emerged as a technique to enhance the interpretability and performance of LLMs by facilitating step-by-step reasoning \cite{wei2022chain}. This method has proven effective in alleviating data bottlenecks and improving model performance on complex tasks, such as mathematical reasoning, commonsense reasoning, and planning \cite{wei2022chain, wang2023self}. CoT allows models to decompose problems into logical intermediate steps, providing greater interpretability compared to traditional end-to-end models \cite{santoro2017simple, xu2024reprompting}. Several techniques have been proposed to enhance CoT reasoning, including few-shot CoT \cite{wei2022chain}, self-consistency \cite{wang2023self}, and Auto-CoT \cite{zhang2023automatic}. Furthermore, strategies like least-to-most prompting \cite{zhou2022least}, AutoHint \cite{sun2023autohint}, and result entropy optimization \cite{wan2023better} have been developed to simplify problem-solving and enhance reasoning accuracy. Specialized approaches like MathPrompter \cite{imani2023mathprompter} for mathematical problem-solving and Meta-prompting \cite{suzgun2024meta} for a refined prompt generation have further expanded the capabilities of CoT.

Recent research, such as THOR \cite{fei2023reasoning}, introduced a CoT reasoning approach to enhance the accuracy and interpretability of sentiment inference through step-by-step reasoning. However, within the realms of fuzzy reasoning and sentiment analysis, the application of CoT is still in its nascent stages. Recent investigations into Zero-Shot-CoT \cite{jin2024zero} have demonstrated that large models can engage in step-by-step reasoning even in the absence of task-specific fine-tuning, presenting a promising avenue for dynamic sentiment inference within a CoT framework. By utilizing CoT, sentiment evaluations can be adjusted dynamically during the inference process, rather than depending solely on static classification systems \cite{dong2024promptexp}. Despite the potential of CoT for fuzzy sentiment analysis, several challenges persist, including the dependence on discrete reasoning and the insufficient integration with continuous fuzzy membership modeling \cite{liu2023pre, liang2018symbolic}.

\section{Fuzzy Reasoning Chain}
In this section, we introduce the Fuzzy Reasoning Chain (FRC) framework, which addresses the unclear reasoning in traditional Chain-of-Thought (CoT) methods and the rigidity of fixed-rule fuzzy reasoning, providing a fuzzy-to-clear inference approach. A comparison between CoT and FRC is shown in Fig.~\ref{FRC}, where FRC refines sentiment analysis through fuzzy membership degrees, facilitating a smoother transition from fuzzy to clear sentiment analysis. Observations of LLMs\cite{guo2025deepseek,liu2024deepseek,qwen2.5,hurst2024gpt} reveal key characteristics in the generated semantic membership degrees, particularly in sentiment analysis:

\textbf{\textit{$\diamond$ Approximate Robustness}}: Small changes in input result in bounded variations in membership degrees when context and syntax remain stable.

\textbf{\textit{$\diamond$ Conditional Monotonicity}}: Sentiment intensity correlates monotonically with membership degrees, except during abrupt context shifts.

\textbf{\textit{$\diamond$ Dynamic Completeness}}: Contextual mechanisms ensure full capture of essential semantic elements and their interactions.

Based on these characteristics, FRC consists of three main steps: Continuous Membership Degree, Multi-Granular Semantic Parsing, and Global Decision Fusion.

\subsection{Continuous Membership Degree}
Let \( C \) represents the sentiment class (e.g., positive or negative), and \( X \) denotes a given text unit. For sentiment analysis, large language models are capable of computing membership degrees reliably, which avoids the need for manually defined membership functions. The membership function \( \mu_C(X) \) can therefore be computed from the output of a large language model as follows:

\begin{equation}
\mu_C(X) = f_{\text{LLM-prompt}}(X, C) \in [0, 1],
\end{equation}

This ensures approximate robustness to input changes, where small variations in the input lead to bounded changes in the membership degree when the context and sentiment class remain stable.

\subsection{Multi-Granular Semantic Parsing}

We expect the large model to use hierarchical reasoning for semantic decomposition and aggregation, ensuring conditional monotonicity. The key steps are:

\textbf{Keyword Membership Degree Calculation:}
We begin by extracting sentiment keywords \( \{k_i\} \) from the input text \( X \) and calculating the corresponding membership degrees for each keyword:
\begin{equation}
\mu_C(k_i) = f_{\text{LLM-prompt}}(k_i, C) \in [0, 1],
\end{equation}
where \( \mu_C(k_i) \) represents the membership degree of keyword \( k_i \) in sentiment class \( C \).

\textbf{Local Semantic Aggregation:}
After extracting the keywords, the next step is to apply a weighted aggregation scheme to the sub-units \( X_j \) of the text. A sub-unit is defined as a portion of the text without emotional overlap, whose membership degree is computed as:
\begin{equation}
\mu_C(X_j) = \max_{k_i \in I_j} \mu_C(k_i),
\end{equation}
where \( I_j \) is the set of keywords \( k_i \) in sub-unit \( X_j \), and the maximum membership degree of the keywords is taken as the membership degree for the sub-unit. This ensures that the strongest sentiment influence from the keywords determines the sentiment of the sub-unit.

\subsection{Global Decision Fusion}

The large language model integrates information from sub-units \( X_j \) to determine the overall sentiment membership degree for input text \( X \). Before the final decision, the model dynamically adjusts the weight based on context, sentiment intensity, and shifts. Key factors considered include, but are not limited to, the following:

\textit{$\diamond$ Language Phenomena:} Shifts in tone, irony, or implication influencing sentiment strength or direction.

\textit{$\diamond$ Sentiment Intensity:} Changes in intensity requiring weight adjustments.

\textit{$\diamond$ Contextual Shifts:} Context changes, e.g., from descriptive to evaluative, triggering adjustments.

This adjustment ensures semantic completeness, capturing relevant elements and interactions for a final sentiment assessment, as follows:

\begin{equation}\label{eq_global}
\mu_C(X) = \sum_{j=1}^m \alpha_{j,C} \cdot \mu_C(X_j),
\end{equation}
where \( m \) is the total number of sub-units \( X_j \), \( \alpha_{j,C} \) is class-specific dynamic weight for sub-unit 
$X_j$ under the class C, satisfying
$\sum_{j=1}^m \alpha_{j,C} = 1$ for each C.

By defining class-specific weight \( \alpha_{j,C} \) for sub-units \( X_j \) and ensuring independence of membership degrees, we can quantify fine-grained sentence emotions and distinguish neutral or conflicting texts, which traditional probability methods cannot do.

\section{Convergence Analysis}
\label{sec:convergence}
This section presents the convergence properties of the FRC framework, focusing on three key characteristics: Approximate Robustness, Conditional Monotonicity, and Dynamic Completeness. These properties contribute to the stability and interpretability of sentiment analysis results, even when the input data is uncertain or imprecise.
\subsection{Approximate Robustness}
\label{sec:robustness}
During the \textit{Keyword Membership Degree Calculation} and \textit{Local Semantic Aggregation} stages, the sentiment membership degree \( \mu_C(X) \) for each subunit is determined based on the strength of sentiment-bearing keywords. Minor input changes, such as syntactic variations or synonym substitutions, lead to smooth, bounded changes in \( \mu_C(X) \), thereby ensuring stable and bounded sentiment analysis.

\textbf{Bounded Analysis:}  
Consider two input texts, \( X \) and \( X' \), with \( d(X, X') \) representing their semantic distance. Assuming that the LLM preserves local smoothness in its semantic mappings, and referencing Eq.\eqref{eq_global}, the difference in membership degrees between \( X \) and \( X' \) can be expressed as:

\[
|\mu_C(X) - \mu_C(X')| = \left| \sum_{j=1}^m g(\alpha_{j,C},X_j) - g(\alpha_{j,C}', X_j') \right|
\]

where $g(\alpha_{C},X) = \alpha_{C} \cdot \mu_C(X)$. Focusing on the difference in membership degrees for each subunit \( X_j \), we recognize that while the output of the language model for each keyword \( k_i \) may not strictly adhere to a continuity constraint, it generally exhibits approximate robustness to small input perturbations. This behavior is especially prominent when the context and syntax of the input remain stable. We hypothesize that the language model demonstrates local stability with respect to minor input variations, so these changes do not cause disproportionately large shifts in the output. Specifically, for each subunit \( X_j \), the difference in its membership degrees relative to \( X_j' \) can be bounded as:
\begin{equation}
    |\mu_C(X_j) - \mu_C(X_j')| \leq L \cdot d(X_j, X_j'),
\end{equation}
where \( L \) is a constant that reflects the model's stability with respect to small input variations, depending on the properties of both the keywords and the context. By summing over all subunits and accounting for the weight adjustments, we derive the final relationship:
\begin{equation}\label{eq-continuity}
    |\mu_C(X) - \mu_C(X')| \leq K \cdot d(X, X'),
\end{equation}
where \( K = \sum_{j=1}^m \alpha_{j,C} \cdot L \). This bounded property enables FRC to exhibit approximate Lipschitz continuity, a phenomenon commonly observed in deep learning models \cite{fazlyab2019efficient, kim2021lipschitz, shang2021lipschitz}, indicating that the FRC framework demonstrates stability and robustness to small perturbations.

\subsection{Conditional Monotonicity}  
\label{sec:monotonicity}

In the \textit{Local Semantic Aggregation} stage, the membership degree \( \mu_C(X_j) \) for each subunit is determined by the sentiment strength \( s(X_j) \) of the relevant keywords. As sentiment strength increases, the membership degree also increases, maintaining a consistent, monotonic relationship. This is preserved in the \textit{Global Decision Fusion}, where the final sentiment membership degree \( \mu_C(X) \) is derived by aggregating the weighted membership degrees of each subunit, as defined in Eq.~\ref{eq_global}.

Given the text \( X \), consider binary sentiment analysis (positive and negative). Traditional Chain-of-Thought (CoT) methods output probabilities, as shown in Fig.~\ref{FRC} (left), where the relationship between sentiment strength and the final output is linearly predictable. Specifically, for a positive sentiment change \( \Delta s_{\text{positive}} \), the probabilities of both positive and negative classes change proportionally:

\[
\begin{aligned}
\Delta P(\text{positive}) &= f_{\text{CoT}}(\Delta s_{\text{positive}}), \\
\Delta P(\text{negative}) &= f_{\text{CoT}}(\Delta s_{\text{positive}}),
\end{aligned}
\]

where \( f_{\text{CoT}} \) is a linear function. The changes are symmetric and linear, ensuring predictable monotonicity.

In contrast, the Fuzzy Reasoning Chain (FRC) independently assigns weights to subunits for each sentiment class, as shown in Fig.~\ref{FRC} (right). The membership degrees for the positive and negative classes \( \mu_{\text{positive}}(X) \) and \( \mu_{\text{negative}}(X) \) are determined separately, which may result in different responses to sentiment changes:

\[
\begin{aligned}
\Delta \mu_{\text{positive}}(X) &= f_{\text{FRC-positive}}(\Delta s_{\text{positive}}), \\
\Delta \mu_{\text{negative}}(X) &= f_{\text{FRC-negative}}(\Delta s_{\text{negative}}).
\end{aligned}
\]

Here, \( f_{\text{FRC-positive}} \) and \( f_{\text{FRC-negative}} \) 
are monotonic mapping functions specific to each sentiment class. As sentiment changes, the impact on the positive and negative classes may differ, leading to less predictable relationships between sentiment intensity and global membership degree. This suggests that FRC's monotonicity might not always follow a linear pattern and could vary across classes.

While CoT typically exhibits a linear, predictable monotonicity, FRC offers more flexibility, potentially allowing for more nuanced relationships between sentiment and class membership. This flexibility could enable FRC to better capture complex sentiment dynamics, although its global monotonicity may not be as straightforward as that of traditional CoT methods

\subsection{Dynamic Completeness}
\label{sec:completeness}
The \textit{Dynamic Completeness} property ensures that all essential semantic elements are captured and appropriately weighted in the final sentiment decision. In FRC, each subunit \( X_j \) represents a critical semantic fragment, and its importance is determined dynamically based on its context. The weights \( \alpha_{j,C} \) are adjusted based on the relevance of each subunit, ensuring that all relevant semantic information contributes to the final membership degree \( \mu_C(X) \).
This is reflected in Eq.\eqref{eq_global}.
By doing so, FRC captures not only individual sentiment influences but also the interactions between these influences, ensuring a comprehensive sentiment analysis.

\subsection{Summary}
Through the analysis of \textit{Approximate Robustness}, \textit{Conditional Monotonicity}, and \textit{Dynamic Completeness}, the FRC framework appears to effectively handle fuzzy inputs, facilitating stable and consistent sentiment analysis. These properties contribute to smooth transitions from uncertainty to clarity, supporting reliable and interpretable results.

\section{Experiments}  

To validate the effectiveness of the Fuzzy Reasoning Chain (FRC) framework in sentiment analysis, we conduct experiments on both English and Chinese datasets. The evaluation focuses on robustness, monotonicity, and classification performance in comparison with CoT-based and direct prompting methods.

\subsection{Datasets}  
In this section, we describe the datasets used in our evaluation. The datasets are divided into two categories: the original datasets used for standard sentiment classification, and the perturbed datasets, which are generated by extending the original datasets for the analysis of FRC convergence and do not require class labels.

\subsubsection{Original Datasets}  
We adopt two well-known datasets for sentiment analysis tasks, selected to ensure both linguistic diversity and domain representation: 

- \textbf{\textit{SemEval-2016 Task 4 (English)}}: A widely used benchmark for sentiment analysis, specifically focused on the evaluation of fine-grained sentiment in social media. It contains a collection of labeled tweets, spanning multiple sentiment classes (5 classes, ranging from negative to positive). The dataset includes over 9,000 labeled tweets, providing a balanced mix of different sentiment categories and domains. The test set consists of 1,791 tweets. 

- \textbf{\textit{Takeout Review Dataset (Chinese)}}: This proprietary dataset is collected from online food delivery reviews and consists of over 10,000 labeled reviews in Chinese. It includes text paired with sentiment labels (positive and negative), focusing on the sentiment expressed in the context of food and restaurant experiences. The large size of this dataset allows for comprehensive training and testing in a real-world application setting.

\subsubsection{Perturbed Datasets}  
To evaluate the robustness and monotonicity of our FRC, we generate perturbed versions of the original datasets using the GPT-4o \cite{hurst2024gpt} API through prompt engineering techniques. These perturbations are designed to test the FRC's ability to handle different levels of changes in the input while preserving core sentiment or adjusting sentiment intensity. The perturbations are divided into two main categories:  

- \textbf{\textit {Robustness Perturbations} }
These perturbations are designed to preserve the original sentiment and meaning while modifying surface-level expressions. Three levels of perturbations are applied:

\textit{Low:} Simple synonym replacements (1-2 words) to test the FRC's resilience to slight lexical changes.  

\textit{Medium:} Sentence restructuring and multiple word replacements to evaluate the FRC's ability to handle moderate perturbations.  
     
\textit{High:} Full sentence rewriting that retains the sentiment but significantly alters sentence structure.  

- \textbf{\textit{Monotonicity Perturbations}}: These perturbations modify sentiment intensity by replacing sentiment-bearing words or adding adverbial modifiers to shift sentiment intensity. The goal is to evaluate the FRC's sensitivity to changes in sentiment intensity, either positive or negative. The labels are used as follows: -1 indicates a more negative sentiment, +1 indicates a more positive sentiment, and 0 indicates no change in sentiment.

- \textbf{\textit{Quality Assurance}}: GPT-4o’s strong performance, combined with manual spot checks, ensures the high quality of the perturbed datasets.


\subsection{Evaluation Metrics}  
To assess the effectiveness of FRC, we propose the following metrics:  

- \textbf{\textit{Robustness Score (RS)}}: Measures the stability of membership degrees or probabilities under perturbations. Given an original text \( X \) and its perturbed counterpart \( X' \), we define:  
    \[
    RS = 1-\frac{1}{N} \sum_{i=1}^{N} \left| \mu_C(X_i) - \mu_C(X'_i) \right|,
    \]
    where \( \mu_C(X) \) is the sentiment membership degree (or probability for CoT) of text \( X \), and $N$ is total test samples. Higher values indicate greater robustness.  

- \textbf{\textit{Monotonicity Score (MS)}}: Evaluates whether increasing sentiment intensity in perturbations leads to a corresponding increase (or decrease) in membership degree. It is computed as:  

    \[
MS = \frac{1}{N} \sum_{i=1}^{N} \mathbb{I}(sgn(\mu_C(X'_i) - \mu_C(X_i))= Y_{i,C}),
\]
where $Y_{i,C}$ is the sentiment shift direction label corresponding to the monotonic perturbation data relative to class $C$ (+1 for more aligned with $C$, -1 for moving away from $C$, 0 for no change), $\mathbb{I}(\cdot)$ is the indicator function (1 if the condition is holds, 0 otherwise). A higher MS value indicates better performance, as it signifies that the model consistently reflects the expected sentiment shift direction.

- \textbf{\textit{F1-score:} } 
Standard metric for sentiment classification.

\subsection{Baseline Methods}  
We compare FRC against two baselines:

- \textbf{\textit{Direct Prompting (DP)}}: Asking LLMs directly for sentiment classification without reasoning steps.  

- \textbf{\textit{CoT}}: A CoT-base method that follows reasoning steps similar to FRC for fairness but outputs probabilities instead of membership degrees. The prompt design is referenced in the left part of Fig.~\ref{FRC}.

\subsection{Experimental Convergence Analysis}  
\textit{\textbf{Theoretical Foundation}}: In Sec.\ref{sec:convergence}, we performed a preliminary convergence analysis of FRC, which indicated that, underpinned by robustness and monotonicity, the FRC framework inherently possesses dynamic completeness.

\textit{\textbf{Experiment Setup}}: Building upon the theoretical foundation, we aim to further investigate the empirical aspects of robustness and monotonicity through targeted experiments. For this purpose, we selected three versions of DeepSeek R1 (32b, 14b, and 7b)\cite{guo2025deepseek} alongside Qwen2.5 32b \cite{qwen2.5} for evaluation. To ensure a more accurate assessment of FRC's convergence, we employed a mixed-language dataset, incorporating both source and perturbed data in Chinese and English, thereby eliminating language as a confounding factor.

\textit{\textbf{Comparison Objective}}: CoT was chosen as a fairness comparison method to highlight FRC's unique advantages in the context of perturbation-induced sentiment shifts. 
\textit{\textbf{Metrics}}: 
The Robustness Score (RS) was used as the evaluation metric, with higher values indicating better performance.
\subsubsection{Robustness Evaluation}
Following the approximate robustness analysis in Sec.\ref{sec:robustness}, we focus on the experimental examination of FRC's robustness. Robustness ensures that small perturbations in input data do not cause significant fluctuations in the output, reinforcing the stability of FRC’s reasoning process. As shown in Table \ref{tab:base}, FRC consistently outperforms CoT across all models and perturbation levels (Low, Medium, High). The average robustness scores reveal a clear performance advantage for FRC. 

For example, DeepSeek-32b achieved an average score of 0.89 with FRC, compared to 0.82 for CoT, resulting in an 8.5\% improvement. Similarly, DeepSeek-14b showed a 10.4\% improvement with FRC (0.85) over CoT (0.77), while Qwen2.5-32b demonstrated an 8.8\% improvement with FRC (0.87) over CoT (0.80).


These results are consistent across models with different parameter sizes, showcasing FRC’s robustness even when scaling down to smaller models like DeepSeek-14b. While the theoretical analysis highlighted FRC’s inherent convergence from fuzziness to clarity, these experimental results further validate its superior robustness, even though the convergence is not absolute. Overall, the findings emphasize FRC’s stability and its ability to handle perturbations that challenge traditional methods like CoT, with particular strength in the robustness of the approximation.

\begin{table}[t]
\renewcommand{\arraystretch}{1.0}
\caption{Robustness Comparison across Models via \textbf{RS} Metric - Higher values indicate better performance.}
\label{tab:base}
\centering
\begin{adjustbox}{width=0.43\textwidth}
\begin{tabular}{lccccr}
\hline
Model & Method & Low & Medium & High & Avg \\
\hline
\multirow{2}{*}{Qwen2.5-32b} & CoT & 0.91 & 0.81 & 0.68 & 0.80 \\
            & FRC & 0.94 & 0.87 & 0.80 & 0.87 \\
\hline
\multirow{2}{*}{DeepSeek-32b} & CoT & 0.92 & 0.83 & 0.72 & 0.82 \\
            & FRC & \textbf{0.95} & \textbf{0.89} & \textbf{0.82} & \textbf{0.89} \\
\hline
\multirow{2}{*}{DeepSeek-14b} & CoT & 0.89 & 0.78 & 0.65 & 0.77 \\
            & FRC & 0.93 & 0.85 & 0.78 & 0.85 \\
\hline
\end{tabular}
\end{adjustbox}
\end{table}

\begin{table}[t]
\renewcommand{\arraystretch}{1.0}
\caption{Monotonicity Evaluation across Models - Higher Monotonicity Score (MS) indicates better performance.}
\label{tab:monotonicity}
\centering
\begin{adjustbox}{width=0.48\textwidth}
\begin{tabular}{lcccc}
\hline
Model & Method & Positive & Negative & Avg MS \\
\hline
\multirow{2}{*}{Qwen2.5-32b} & CoT & 0.84 & 0.82 & 0.83 \\
            & FRC & 0.93 & 0.89 & 0.91 \\
\hline
\multirow{2}{*}{DeepSeek-32b} & CoT & 0.86 & 0.84 & 0.85 \\
            & FRC & \textbf{0.94} & \textbf{0.90} & \textbf{0.92} \\
\hline
\multirow{2}{*}{DeepSeek-14b} & CoT & 0.81 & 0.79 & 0.80 \\
            & FRC & 0.91 & 0.87 & 0.89 \\
\hline
\end{tabular}
\end{adjustbox}
\end{table}

\subsubsection{Monotonicity Evaluation}

In addition to robustness, monotonicity is another key characteristic that influences FRC’s convergence properties. This section investigates how the sentiment intensity of input perturbations correlates monotonically with the changes in membership degrees, and how FRC leverages this relationship to refine its reasoning and sentiment analysis.

The results for monotonicity evaluation across models are provided in Table \ref{tab:monotonicity}. FRC demonstrates superior performance in capturing both positive and negative sentiment shifts, as shown by higher average MS scores across all models. For example, DeepSeek-32b with FRC achieved an average MS of 0.92, compared to 0.85 for CoT, indicating a significant improvement in the model's ability to reflect sentiment changes. Similarly, DeepSeek-14b and Qwen2.5-32b showed similar gains in MS with FRC.

\subsection{Sentiment Classification}  
We evaluate DP, CoT, and FRC on the SemEval-2016 Task 4 and Takeout Review datasets, which provide positive and negative labels, with Task 4 also including neutral cases. Some samples exhibit inherently ambiguous sentiment, leading conventional prompt-based methods to produce unstable or neutral predictions. Using FRC’s coarse positive/negative assessment, we observed that the membership degree differences for these samples are typically within 0.3. Based on this, we divide the data into \textit{Clear} cases, with differences above 0.3, and \textit{Ambiguous} cases, with differences at or below 0.3. This allows us to retain the original labels while enabling more detailed analysis. Even without relying on the ground-truth labels, finer distinctions, such as strong versus weak sentiment conflicts, can be examined by comparing membership degrees. For evaluation, the predicted label is assigned according to the higher membership degree, with equal values considered neutral. The \textit{Average} category then reports overall performance across both \textit{Clear} and \textit{Ambiguous} cases.

\begin{table}[ht]
\renewcommand{\arraystretch}{1}
\caption{Sentiment Classification Results on SemEval-2016 Task 4 (English) using F1 Score, divided into Clear, Ambiguous, and Average categories.}
\label{tab:semeval}
\centering
\begin{adjustbox}{width=0.48\textwidth}
\begin{tabular}{lcccc}
\hline
Model & Method & Clear & Ambiguous & Avg \\
\hline
\multirow{3}{*}{Qwen2.5-32b}  & DP & 0.86 & 0.81 & 0.84 \\
             & CoT & 0.88 & 0.82 & 0.85 \\
             & FRC & \textbf{0.90} & \textbf{0.85} & \textbf{0.88} \\
\hline
\multirow{3}{*}{DeepSeek-32b} & DP & 0.83 & 0.75 & 0.79 \\
             & CoT & 0.85 & 0.77 & 0.81 \\
             & FRC & \textbf{0.89} & \textbf{0.84} & \textbf{0.87} \\
\hline
\multirow{3}{*}{DeepSeek-14b} & DP & 0.77 & 0.72 & 0.75 \\
             & CoT & 0.79 & 0.73 & 0.76 \\
             & FRC & \textbf{0.83} & \textbf{0.78} & \textbf{0.81} \\

\hline
\end{tabular}
\end{adjustbox}
\end{table}

\begin{table}[ht]
\renewcommand{\arraystretch}{1}
\caption{Sentiment Classification Results on Takeout Review Dataset (Chinese) using F1 Score, divided into Clear, Ambiguous, and Average categories.}
\label{tab:takeout}
\centering
\begin{adjustbox}{width=0.48\textwidth}
\begin{tabular}{lcccc}
\hline
Model & Method & Clear & Ambiguous & Avg  \\
\hline
\multirow{3}{*}{Qwen2.5-32b}  & DP & 0.84 & 0.79 & 0.81 \\
             & CoT & 0.86 & 0.80 & 0.83 \\
             & FRC & \textbf{0.88} & \textbf{0.83} & \textbf{0.86} \\
\hline
\multirow{3}{*}{DeepSeek-32b} & DP & 0.81 & 0.74 & 0.77 \\
             & CoT & 0.84 & 0.76 & 0.80 \\
             & FRC & \textbf{0.89} & \textbf{0.84} & \textbf{0.87} \\
\hline
\multirow{3}{*}{DeepSeek-14b} & DP & 0.73 & 0.68 & 0.70 \\
             & CoT & 0.75 & 0.71 & 0.74 \\
             & FRC & \textbf{0.79} & \textbf{0.74} & \textbf{0.77} \\
\hline
\end{tabular}
\end{adjustbox}
\end{table}
The experimental results demonstrate that FRC consistently outperforms both DP and CoT across all models and datasets. 

\textbf{\textit{SemEval-2016 Task 4}}: As shown in Table \ref{tab:semeval}. , FRC achieves the highest F1 scores in both the \textit{Clear} and \textit{Average} categories, showing substantial improvements over CoT and DP. 

\textbf{\textit{Takeout Review Dataset}}:
On the Takeout Review Dataset (Chinese), as shown in Table \ref{tab:takeout}, FRC also outperforms the baseline methods in both \textit{Clear} and \textit{Average} categories, indicating its superior ability to distinguish sentiment in diverse languages.

The results show that the difference in performance between FRC and the other methods is particularly pronounced in the \textit{Clear} category, where FRC demonstrates more robust sentiment classification. This is consistent with FRC’s ability to process sentiment shifts and handle fine-grained sentiment distinctions more effectively.

Furthermore, the model size appears to have a positive impact on FRC’s performance, with larger models such as DeepSeek-32b and Qwen2.5-32b consistently achieving better performance across both datasets.

\subsection{Knowledge Transfer from Large to Small Models}  

To evaluate the knowledge transfer capability of the FRC framework from large to small models, we conduct prompt-based experiments by injecting intermediate reasoning results, specifically keyword knowledge and sub-unit knowledge, extracted from the DeepSeek-R1 32b model into the prompts of smaller models, including the 7b and 1.5b models. This strategy, similar to retrieval-augmented generation, provides structural reasoning guidance without modifying any model parameters.
As shown in Table~\ref{tab:transfer}, the injected knowledge leads to substantial improvements in the 7b model’s F1-score.


\begin{table}[ht]
\centering

\caption{Knowledge Transfer from DeepSeek-R1 32b to Smaller Models via FRC (F1-Score Results)}
\begin{adjustbox}{width=0.45\textwidth}
\begin{tabular}{ccc}
\hline
\textbf{Prompt Configuration} & \textbf{1.5b}& \textbf{7b}\\
\hline
No injection (baseline) & 0.62& 0.76 \\
Injecting Keyword Knowledge & 0.68 & 0.79 \\
Injecting Sub-unit Knowledge & 0.72 & 0.81 \\
Injecting Keyword \& Sub-unit Knowledge & \textbf{0.75} & \textbf{0.83}\\

\hline
\end{tabular}
\end{adjustbox}

\label{tab:transfer}
\end{table}

\textbf{Analysis:}  
The baseline 1.5b and 7b models achieve F1-scores of 0.62 and 0.76, which are substantially lower than the 32b model, which reaches 0.87. This gap indicates the limitations of smaller models in handling long prompts, maintaining global context, and performing complex reasoning. By injecting fine-grained keyword and sub-unit knowledge derived from the 32b model, these deficiencies are effectively mitigated. In particular, the 1.5b model reaches 0.75 F1-score, corresponding to a relative improvement of 21 percent, while the 7b model achieves 0.83 F1-score. These results demonstrate that smaller models can narrow the performance gap without any parameter updates.

The improvements obtained by injecting keyword knowledge alone, sub-unit knowledge alone, and their combination indicate that multi-level knowledge components complement each other in enhancing reasoning performance. Moreover, this offline knowledge transfer mechanism enables small models to process new sentences independently by querying a pre-built knowledge base or leveraging offline-enhanced training, without requiring online invocation of large models. This approach allows for efficient and low-cost deployment while maintaining robust reasoning capability.

\section{Conclusion}

In this work, we propose the Fuzzy Reasoning Chain (FRC) framework to address the challenge of reasoning over ambiguous and uncertain texts. FRC extends conventional probability-based reasoning by integrating continuous fuzzy membership degrees, providing a structured mechanism for translating fuzzy information into interpretable decisions. We provide a preliminary exploration of the framework’s properties, and empirical results demonstrate its effectiveness on sentiment analysis tasks. The transition from probabilities to membership degrees allows FRC to capture conflicting or uncertain signals that traditional approaches cannot, and also facilitates knowledge transfer between models of different scales. Future work may explore applying FRC to other fuzzy reasoning tasks such as security review and ethical judgment, integrating richer external knowledge sources, and further improving robustness and efficiency for practical deployments.


\section{Limitations} 
While the proposed Fuzzy Reasoning Chain (FRC) framework shows promising results, several limitations remain to be addressed in future work.

First, current large language models still face challenges in precisely executing the FRC prompting process end-to-end. Similar to many existing prompt-based tasks\cite{kojima2022large}, practical deployment often requires manual interpretation of intermediate outputs or multiple rounds of interaction with the model , which may limit efficiency and scalability.

Second, although we focus on sentiment analysis for evaluation, other fuzzy reasoning tasks—such as security review, ethical judgment, and culturally sensitive scenarios—pose additional complexities due to their context-dependent semantics and domain-specific knowledge requirements. Careful task-specific adaptation and thorough empirical validation are needed before applying FRC to these areas.

Third, the generalization of FRC beyond sentiment analysis remains to be systematically explored. Different fuzzy tasks may have distinct ambiguity types and semantic structures, which could impact the framework’s reasoning stability and interpretability.

Fourth, the reliance on large language models as semantic priors implies a dependency on the inherent biases and knowledge limitations of these models. This may affect the robustness of FRC’s reasoning in scenarios with underrepresented or evolving language phenomena.

Finally, while FRC facilitates knowledge transfer via prompt injection, the granularity and optimal design of transferable reasoning components warrant deeper investigation to maximize cross-model efficacy.

Addressing these limitations will be crucial for further enhancing FRC’s applicability and robustness in diverse fuzzy reasoning domains.

\section{Ethics and Impact Statement}

This paper presents the Fuzzy Reasoning Chain (FRC) framework designed to improve the interpretability and robustness of sentiment analysis through fuzzy logic-based reasoning. By enabling models to handle ambiguous or uncertain emotional expressions more effectively, our work aims to enhance natural language understanding in real-world scenarios.

While we do not identify immediate ethical concerns inherent to our methodology, we acknowledge that sentiment analysis technologies can influence user experience and decision-making in various applications such as content moderation, recommendation systems, and social media analysis. It is therefore important that future work applying FRC continues to consider fairness, privacy, and the avoidance of biased or misleading interpretations.

Our goal is to support the development of more nuanced and reliable sentiment analysis tools that better reflect human emotional complexity, ultimately benefiting users and society.

\bibliography{custom}




\end{document}